\documentclass{article}

\usepackage{arxiv}

\usepackage[utf8]{inputenc} 
\usepackage[T1]{fontenc}    
\usepackage{hyperref}       
\usepackage{url}            
\usepackage{booktabs}       
\usepackage{amsfonts}       
\usepackage{nicefrac}       
\usepackage{microtype}      
\usepackage{lipsum}		
\usepackage{graphicx}
\usepackage{natbib}
\usepackage{doi}

\title{Non-Invasive Qualitative Vibration Analysis using Event Camera}

\author{\href{https://orcid.org/0009-0005-3423-8492}{\includegraphics[scale=0.06]{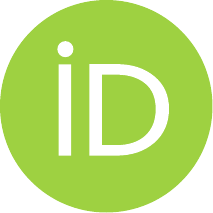}\hspace{1mm}Dwijay Bane$^{+}$},\hspace{1mm}Anurag Gupta, \href{https://orcid.org/0000-0001-5089-2658}{\includegraphics[scale=0.06]{orcid.pdf}\hspace{1mm}Manan Suri$^{+}$}
        \\
	Department of Electrical Engineering\\
	Indian Institute of Technology Delhi\\
	New Delhi, India - 110016 \\
	\texttt{eey207529@ee.iitd.ac.in, manansuri@ee.iitd.ac.in} \\
}

\renewcommand{\undertitle}{Technical Report}

\hypersetup{
pdftitle={Non-Invasive Qualitative Vibration Analysis using Event Camera},
pdfsubject={cs.NE},
pdfauthor={Dwijay Bane, Anurag Gupta, ...., Manan Suri},
pdfkeywords={Event Camera, Micro-Vibration, Motion Magnification},
}

\begin{document}
\maketitle

\begin{abstract}
This technical report investigates the application of event-based vision sensors in non-invasive qualitative vibration analysis, with a particular focus on frequency measurement and motion magnification. Event cameras, with their high temporal resolution and dynamic range, offer promising capabilities for real-time structural assessment and subtle motion analysis. Our study employs cutting-edge event-based vision techniques to explore real-world scenarios in frequency measurement in vibrational analysis and intensity reconstruction for motion magnification. In the former, event-based sensors demonstrated significant potential for real-time structural assessment. However, our work in motion magnification revealed considerable challenges, particularly in scenarios involving stationary cameras and isolated motion.
\end{abstract}

\keywords{Event Camera \and Micro-Vibration \and Motion Magnification}

\section{Introduction}
\par Event cameras represent a revolutionary approach to visual sensing, offering significant advantages over conventional frame-based imaging technologies. Unlike traditional cameras that capture absolute light intensity values, event cameras detect and record changes in light intensity, resulting in exceptional motion blur resistance, high temporal resolution, and reduced power consumption\citep{9129849} \citep{9138762}. Multiple techniques have been adopted to make use of these qualities for object tracking, classification, gesture recognition, etc. \citep{8593805} \citep{10.1145/3477145.3477263} \citep{8100264} \citep{8702189} \citep{10.3389/fnins.2023.1149410}. This technical report explores the qualitative analysis of event camera technology, focusing on two advanced applications: capturing high-precision micro-vibration and motion magnification. Section~\ref{sec:background} discusses background and literature survey. Section~\ref{sec:ExperimentalSetup} refers to experimental setup. Section~\ref{sec:ExperimentsResults_EBVA} evaluates qualitative results and section~\ref{sec:conclusion} mentions the conclusion and future scope.

\section{Background}
\label{sec:background}

\subsection{Event Camera and Data Format}
\begin{figure}[h]
\centerline{\includegraphics[width=\linewidth]{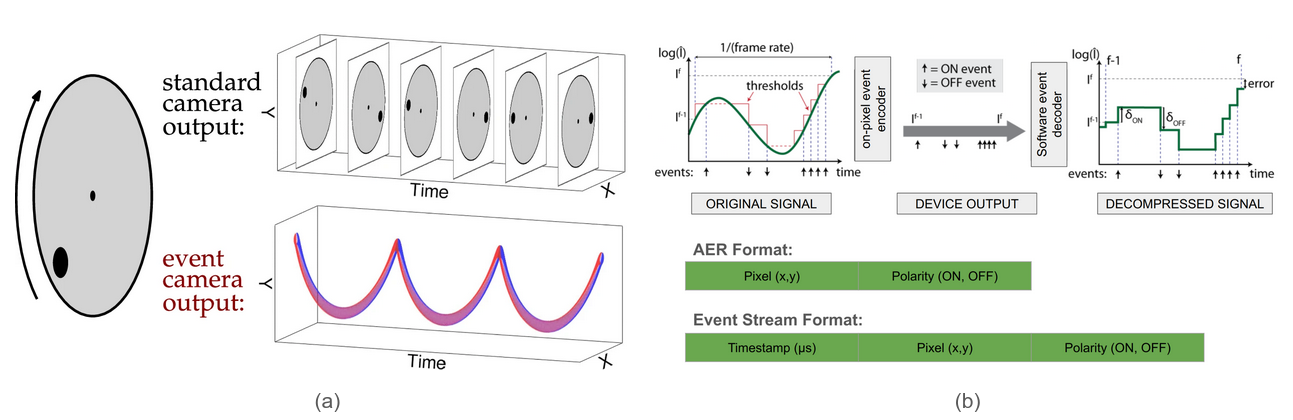}}
\caption{Frame based vs Event Camera, (a.) shows how both sensors capture data of spinning disc and (b.) shows Event Sensor AER encoding, decoding and format. From \citep{inbookDavide} \citep{6865228}.}
\label{fig:eventConti}
\end{figure}

\par Event cameras are engineered to mimic the human eye's functionality, responding solely to changes in brightness within their field of view. This dynamic sensing method eliminates the concept of fixed frame rates, instead converting visual information into sequences of spikes or events. By doing so, event cameras create multiple encodings of the visual scene, translating the spatio-temporal patterns of incident light into discrete, asynchronous signals that represent changes in the environment. Figure \ref{fig:eventConti} ((a.) bottom) illustrates this concept by presenting events plotted in a 3D space-time representation. This visualization reveals how different types of motion generate distinct event patterns. For instance, a rotating disc creates a spiral of events in space-time, while moving objects trigger events along their edges. Event cameras transmit data from their pixel array to external systems using a shared digital output bus, commonly employing the address-event representation (AER) protocol (\citep{1310498}, \citep{OnChipAER}, \citep{842110}). An AER (see Fig:\ref{fig:eventConti} (b.)) event consists of polarity (ON/OFF) and pixel address (x, y) clubbed with a microsecond timestamp in event stream format. The event data format from an event camera is represented as a tuple \(e = (x, y, p, t)\), where an event with polarity \(p\in\{-1,1\}\) is generated at the pixel coordinate \( (x, y) \) at timestamp \(t\).

\subsection{Frequency Estimation Basics}
\par Frequency measurement is the process of identifying repeated events within a specific time frame, often used with signal or event-based data. It’s crucial for spotting temporal patterns or periodicity in event-based applications, which aids in detecting dynamic changes, recognizing motion differences, or identifying anomalies. By quantifying the frequency content at specific spatial locations, it characterizes dynamic processes, making it applicable in motion analysis, vibration monitoring, or event-driven systems evaluations.

\par Research on frequency detection using event-based cameras are catching traction \citep{6696456} \citep{8516629} \citep{pfrommer2022frequencycamimagingperiodic}. An early notable contribution to this field is found in \citep{6696456}, which focuses on detecting blinking LED markers. The authors present an algorithm designed to identify the presence of specific frequencies within a signal from a predetermined set. Their approach, centered on detecting signal peaks through polarity transitions. Vibrational frequency is usually measured using event data with laser-assisted illumination, as described in \citep{10208228}. Further analysis is supported by spiking neural networks (SNN) \citep{10016699}. This ability to detect micro-vibrations is crucial for Fault Diagnosis applications, including bearing fault detection \citep{s21124070}, multi-sensor fault diagnosis \citep{9715938}, and delamination detection in composite plates \citep{s19071734}.

\subsection{Reconstruction of Intensity Image from Event Stream}
\label{sec:ReconstructBasics}

\begin{figure}[h]
\centerline{\includegraphics[scale=0.8]{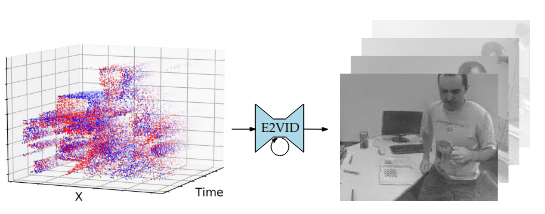}}
\caption{Shows event data in 3D (x,y,t) voxel grid on left and its reconstructed intensity image on right. Adapted from \citep{rebecq2019highspeedhighdynamic}.}
\label{fig:reconstruct_e2vid}
\end{figure}

\par In theory, event streams contain all the necessary information to fully represent a visual signal. However, in practice, reconstructing an intensity image from these event streams poses a significant challenge, making it an ill-posed problem. Current reconstruction methods rely on manually designed priors and make significant assumptions regarding both the imaging process and the statistical properties of natural images.
\newline The E2VID method \citep{rebecq2019highspeedhighdynamic} presents an innovative approach that utilizes machine learning techniques to directly transform event streams into intensity images, eschewing the need for hand-crafted priors. At the core of this method is an innovative recurrent neural network architecture designed specifically for video reconstruction from event data. To train this network effectively, the authors employed an extensive collection of synthetic event data to train their model, allowing it to discern the intricate correlations between event streams and their associated intensity images (see Fig:\ref{fig:reconstruct_e2vid}). The purpose of reconstructing Intensity Image is to get image without motion blur, benefits of wide dynamic resolution of event camera on which off-the-shelf algorithms can be implemented.

\subsection{Motion Magnification Basics}
\par Motion magnification is an advanced computational technique that amplifies subtle movements in video footage, rendering them more perceptible to the human eye. This powerful method has found widespread applications across various fields where the analysis and monitoring of minute motions are crucial. The process employs sophisticated algorithms to detect and amplify slight changes in pixel intensity or motion patterns between successive video frames. By emphasizing these subtle variations, motion magnification enables researchers and analysts to visualize and study movements that would otherwise remain imperceptible.

\par Lagrangian Approaches: \newline Early motion magnification techniques \citep{MotionMagLiu} \citep{MotionMagWang}, utilized Lagrangian methods, relying on explicit motion computation and frame warping. Although these methods were effective, they struggled with accurate motion estimation, resulting in noticeable errors in the magnified outputs. Additionally, their high computational demands limited practical applications.

\par Eulerian approaches: \newline To overcome the limitations of Lagrangian methods, Eulerian approaches were introduced as a more efficient alternative. \citep{EulerianFuchs} and \citep{Wu12Eulerian} developed Eulerian video processing techniques that bypassed the need for computationally expensive flow calculations. However, linear Eulerian methods had their own drawbacks, allowing only small magnification at high spatial frequencies and amplifying noise as magnification increased.

\subsection{Complex-Valued Steerable Pyramids Technique for Motion Magnification}
\label{sec:steerablepyr}
\par This work introduces a novel Eulerian approach based on complex-valued steerable pyramids, expanding on the foundational research of \citep{SteerablePyrSimoncelli} and \citep{SteerablePyrPortilla2000}. The method is inspired by phase-based optical flow \citep{FleetJepson} and the concept of motion without movement \citep{FreemanMotionWithoutMovement}. By leveraging complex-valued steerable pyramids, this approach provides an efficient framework for magnifying subtle motions without the need for explicit optical flow calculations.

\noindent The technique exploits the relationship between motion and phase in steerable pyramids. Instead of directly computing motion, it measures local motion by analyzing phase variations in spatial subbands. These phase shifts correspond to small, often imperceptible movements, which the method can then amplify. This phase-based strategy enables precise motion magnification while bypassing the computational intensity typically associated with optical flow techniques.

\noindent To overcome spatial limitations inherent in steerable basis functions, the approach extends the steerable pyramid to sub-octave bandwidth pyramids. Although this extension increases the redundancy of the representation, it allows for more substantial motion amplification across all spatial frequencies. As a result, the method produces fewer artifacts and demonstrates improved noise performance compared to traditional linear Eulerian methods.

\noindent This method involves transforming each video frame into complex pyramids and then applying Fast Fourier Transforms (FFTs) to identify which components of the image to magnify \cite{10.1145/2461912.2461966}. 

\begin{figure}[h]
\centerline{\includegraphics[width=\linewidth]{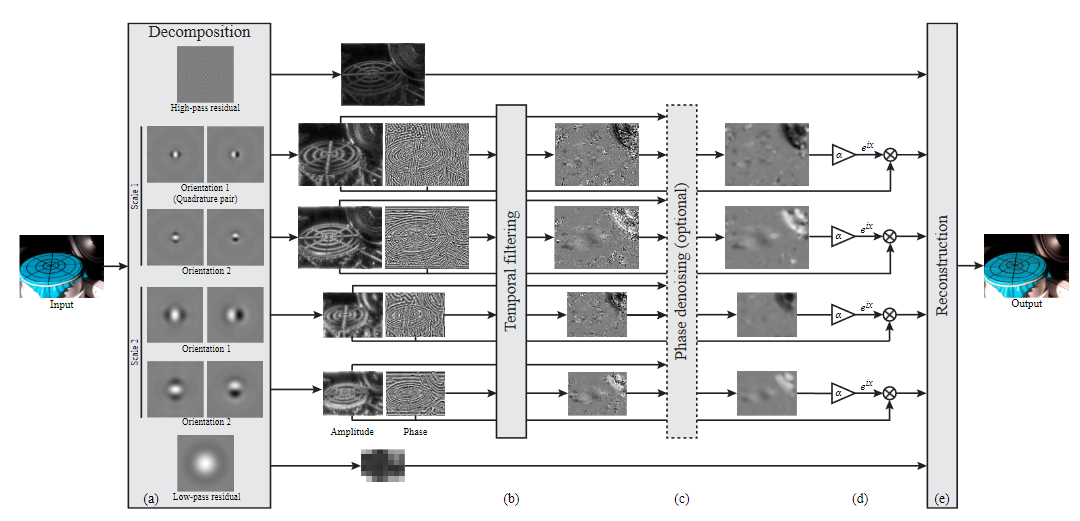}}
\caption{(a) Frame decomposition using steerable pyramids, (b) Temporal filtering to identify phases in the frequency range we want to magnify, (c) Phase denoising, (d) Magnifying the filtered phases, (e) Reconstructing the frame}
\label{fig:motionMag}
\end{figure}

\noindent In continuous space, a frame can be represented as:
\begin{equation}
   f(x) = \displaystyle\sum_{k=-\infty}^{\infty} c_k \exp(j2\pi kx) \label{eq:frame}
\end{equation}
where \(c_x\) is some complex number
\newline From Fourier shift theorem, we can say that
\begin{equation}
   f(x + \delta(t)) = \displaystyle\sum_{k=-\infty}^{\infty} c_k \exp(j2\pi k(x + \delta(t))) \label{eq:frameWithFourier}
\end{equation}
Here \(\delta(t)\) represents the slight motion at a position x over time.
\begin{equation}
   \Delta\phi(t) = 2\pi k\delta(t) \label{eq:frameIter}
\end{equation}
For motion magnification, we choose a particular frequency range in which we want to magnify the phase \(\delta(t)\). This can be done by temporal filters, either Butterworth or FIR. 
\newline Thus, \(\Delta\phi(t)\) is the changing phase which we want to amplify. Taking magnification factor, m (>1 for magnification, <1 for attenuation),
\begin{equation}
   f_{mag}(x + \delta(t)) = \displaystyle\sum_{k=-\infty}^{\infty} c_k \exp(j2\pi k(x + \delta(t))).\exp(jm\Delta\phi(t)) \label{eq:frameWithMag}
\end{equation}
This can be converted back to an image frame and joined together to get back the video with magnified motion. Fig.~\ref{fig:motionMag} highlights the basic pipeline for motion magnification.

\section{Experimental Setup}
\label{sec:ExperimentalSetup}
\subsection{Hardware Setup}
\begin{figure}[h]
\centerline{\includegraphics[scale=0.6]{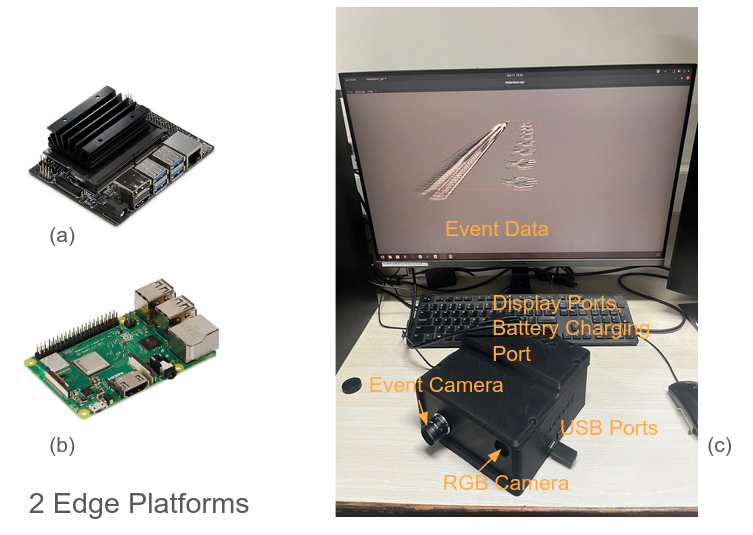}}
\caption{Hardware Setup for experimentation. It facilitates integration of two Edge Platforms. (a) NVIDIA Jetson Nano (b) Raspberry Pi 3B+ and (c) complete setup with connections.}
\label{fig:NewSetup}
\end{figure}

In our setup we created a case for capturing event data from event based vision sensor and frame based data from a CSI/MIPI interfaced camera in a stereo configuration over a edge platform, 2 platforms are currently supported which includes Raspberry Pi 3B+, for this work NVIDIA Jetson Nano (1.4 GHz ARM Cortex-A57) is used with NVIDIA Tegra-X1 GM20B GPU for graphics and CUDA processing which is efficient edge compute performance. Fig:\ref{fig:NewSetup} shows our setup while capturing event data of hand movement in 3D (x,y,t) plot visualizer where both event stream and frames are shown side-by-side. 

\noindent In our setup Prophesee EVK-4 event sensor is interfaced with USB-3 protocol to support higher event data readout upto 1Geps(Giga events per second). Here EVK-4 uses a C/CS-Mount 8mm manual focus optical lens with Focal Length adjustment range from F2 to F11. The reason EVK-4 sensor is used due to its temporal resoultion of 1 $\mu$s per pixel (equivalent to >10k FPS) and no threshold on output buffer, it can capture as much recording device have storage.

\noindent In the right side RGB sensor placeholder, Raspberry-Pi RGB Camera v2.1 is used which has Sony IMX219 8-megapixel sensor. It attaches via a 15cm ribbon cable to the CSI port on the NVIDIA Jetson Nano. 

\subsection{Bias Tuning of Event Sensors}
\label{sec:biasTuning}
\begin{table}[tb]
\begin{center}
\begin{tabular}{|l|l|l|l|}
\hline
\multicolumn{1}{|c|}{\textbf{Bias Name}} & \multicolumn{1}{c|}{\textbf{Value Used}} & \multicolumn{1}{c|}{\textbf{Minimum value}} & \multicolumn{1}{c|}{\textbf{Maximum value}} \\ \hline
bias\_diff                               & 0                                        & -25                                         & 23                                          \\ \hline
bias\_diff\_on                           & -15                                      & -85                                         & 140                                         \\ \hline
bias\_diff\_off                          & -15                                      & -35                                         & 190                                         \\ \hline
bias\_fo                                 & 10                                       & -35                                         & 55                                          \\ \hline
bias\_hpf                                & 0                                        & 0                                           & 120                                         \\ \hline
bias\_refr                               & 0                                        & -20                                         & 235                                         \\ \hline
\end{tabular}
\caption{Bias Tuning for EVK-4 sensor in indoor settings}
\label{tab:biastuning}
\end{center}
\end{table}

\noindent EVK-4 features customizable sensor parameters known as 'biases'. These biases allow users to fine-tune the sensor's performance to meet specific application needs and environmental conditions. By adjusting these settings, users can optimize various aspects of the sensor's operation, such as increasing temporal resolution for high-speed applications, reducing background noise in the event stream and modifying contrast sensitivity thresholds. This flexibility enables the sensor to adapt to a wide range of use cases, enhancing its versatility across different scenarios.

\noindent Type of bias tuning\citep{propheseeBiasesx2014} that is possible in EVK-4:
\begin{itemize}
\item \textbf{Contrast Sensitivity Bias:} \textit{bias\_diff\_on} and \textit{bias\_diff\_off} controls the pixel sensitivity to positive and negative light changes respectively. When both contrast sensitivity thresholds are decreased, it consequently improves pixel sensitivity, improves pixel latency, increases background rate and also can generate "stuck-at" pixels. When both contrast sensitivity thresholds are increased, it then consequently reduces pixel sensitivity, deteriorates pixel latency, decreases background rate and also reduces the number of generated events.
\item \textbf{Low-Pass and High-Pass Filters:} \textit{bias\_fo} and \textit{bias\_hpf} adjusts the low pass and high pass filters respectively. Decreasing low-pass filter smooths the signal out removing high frequencies and increases pixel latency while increasing low-pass filter improves pixel latency and also increases noise, even rate and range of observed high-frequencies. Decreasing high-pass filter preserves low frequency signal, slow motion and increases noise while increasing high-pass removes low-frequency signal, slow motion and decreases noise.
\item \textbf{Refractory Period Bias:} \textit{bias\_refr} controls the pixel "refractory (dead) period". Decreasing the pixel refractory period improved pixel availability, increases the number of events triggered by contrast change and helps signal tracking while increasing dead time reduces the number of events triggered by contrast change and can also used to remove one polarity for periodic high frequency signal.
\end{itemize}

\noindent Table~\ref{tab:biastuning} shows bias tuning values used for this work. Applications in this work are mostly recorded in indoor room or semi-open room. So bias tuning will change in case the data capture needs to be done in noisy complex environment.

\subsection{Data Pre-processing}
To further tune event data on software level, prophesee's event signal processing modules\citep{propheseeEventSignal} for event data filtering is applied. Out of 3 modules which are Anti-Flicker (AFK), Spatio-Temporal-Contrast (STC), Event Rate Control (ERC), we have mostly used the combination of STC filter and ERC. STC filter keeps the second event in a burst and filters out isolated events, with two variants allowing either removal or retention of subsequent events in the trail. ERC mechanism dynamically adjusts the spatial and temporal distribution of events. It ensures the event rate remains within the system's maximum processing capacity, optimizing performance and preventing data overflow. Also most importantly, when the ERC engages, it may result in reduced signal quality.

\noindent For decoding and processing event in python library we have used OpenEB\citep{propheseeOpenEB} which is provided by prophesee under open source license. We have used Anaconda\citep{anaconda} open-source python distribution python-3.10 as python environment in ubuntu 20.04. For NVIDIA Jetson Nano Jetpack-SDK with open-source drivers were used to interface EVK4 device.

\section{Experiments and Results}
\label{sec:ExperimentsResults_EBVA}

\subsection{Qualitative Evaluation of Frequency Visualization}
Our method to find frequency was by calculating the time between successive hypertransitions. Hypertransitions represent the time intervals between polarity changes in event-based data. The polarity changes occur when individual pixels or sensors in the event camera detect a change in brightness or intensity. These changes are captured as events containing information about their spatial location \((x, y)\), polarity (on/off), and timestamp \(t\).

\begin{figure}[h]
\centerline{\includegraphics[width=\linewidth]{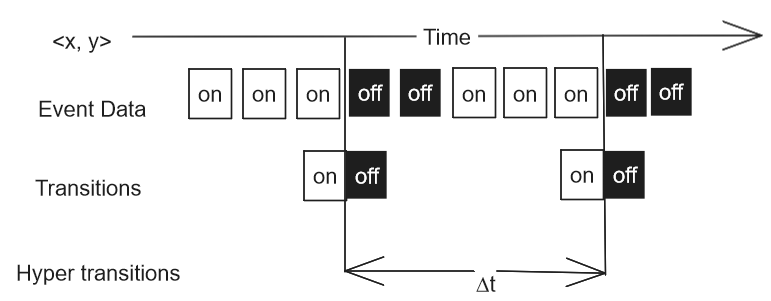}}
\caption{Hyper transition in event data. Adapted from \citep{6696456}}
\label{fig:edtransition}
\end{figure}

\noindent To calculate hypertransitions at a particular spatial location \((x, y)\):
\newline \textbf{\textit{Polarity Change Detection:}} Identify consecutive transitions at location \((x, y)\) when the polarity changes from on to off (or vice versa).
\newline \textbf{\textit{Compute Time Intervals:}} Calculate the time intervals between consecutive transitions. These intervals represent the hypertransitions at that spatial location.
\newline \textbf{\textit{Mean Hypertransition:}} Thus frequency estimation at the point \((x, y)\) is then derived by analysing these hypertransitions, typically by computing the mean or reciprocal of the mean hypertransition duration \(\Delta t\) (see Fig.~\ref{fig:edtransition}).

In this study, we focus on qualitative analysis of event-based vision techniques for real-time vibration monitoring and predictive maintenance in real-world applications. 
Our work uses hypertransition from \citep{6696456}, which is from ON to OFF events, a choice that our observation confirms as having reduced jitter and noise compared to OFF to ON transitions and observed 0.3 FPS with 3 sec for processing >15 ms temporal batch. Previous work \citep{8516629} on OFF to ON transition uses repeatability of 16 transition before estimating frequency to reduce noise and jitter but increases delay before estimation. The underlying cause of rising edge transition being less stable likely stems from the behavior of the pixel's photo-current prior to the first ON event. As mentioned for the DVS sensor in section 3.1 of \citep{hu2021v2evideoframesrealistic}, the photo-current drops to very low levels before an ON event occurs, leading to extended response times. \citep{pfrommer2022frequencycamimagingperiodic} further demonstrates the use of OFF to ON transition with IIR filter for reducing noise and also interpolating frequency using half period at zero-crossing giving better accuracy. We have adapted their efficient state based processing and visualization from their opensource library of FrequencyCam\citep{pfrommer2022frequencycamimagingperiodic} made for ROS. Their approach increased latency to 10 sec for single frame which is 0.1 FPS when considering >15 ms temporal batches. Consequently, for the EVK4 with resolution of 1280x720 pixels, we decided to use ON to OFF transition giving better balance of stable transition with lesser accuracy while offering 3x higher FPS for observing per-pixel frequency map with reduced jitter and noise. There is significant role of tweaking the bias tuning mentioned in Section~\ref{sec:biasTuning} for keeping event camera sensitive to faster change and have reduced noise.

\noindent Given the sampling frequency of EVK4\citep{prop:prophesee} which is 10 kHz. Therefore, theoretically, with a 100$\mu$s latency, an event camera could capture frequencies up to 5 kHz without aliasing. In all our data capture we used 8mm C-Mount optical lens, which was configured on Focal length of F4 or F8. Table~\ref{tab:evk4} shows used datasets in our experimentation, where 2 sample dataset were taken from prophesee public datasets\citep{propheseeRecordingsDatasets} and rest samples were recorded internally with high resolution (1280x720) EVK-4 sensor.

\begin{table}[tb]
    \begin{center}
    \begin{tabular}{|c|c|c|c|c|}
    \hline
    \textbf{No.} & \textbf{Name} & \textbf{Duration} & \textbf{No. of} & \textbf{Event Sensor} \\
     & & &\textbf{recordings}&\\
    \hline
    1 & monitoring\_40\_50hz\citep{propheseeRecordingsDatasets}& 6sec & 1 & Prophesee Gen-3.0\\
    \hline
    2 & hand\_spinner\citep{propheseeRecordingsDatasets}& 5sec & 1 &  Prophesee Gen-3.1\\
    \hline
    3 & Light Sources& 35sec & 4 &  EVK-4\\
    \hline
    4 & Pedestal Fan& 10min & 2 &  EVK-4\\
     \hline
    5 & Water pumping station& 1.2min & 8 &  EVK-4\\
     \hline
    \end{tabular}
    \end{center}
\caption{Dataset summary showing public samples (No. 1,2) and custom recorded samples (No. 3,4,5).}
\label{tab:evk4}
\end{table}

\begin{figure}[h]
\centerline{\includegraphics[width=\linewidth]{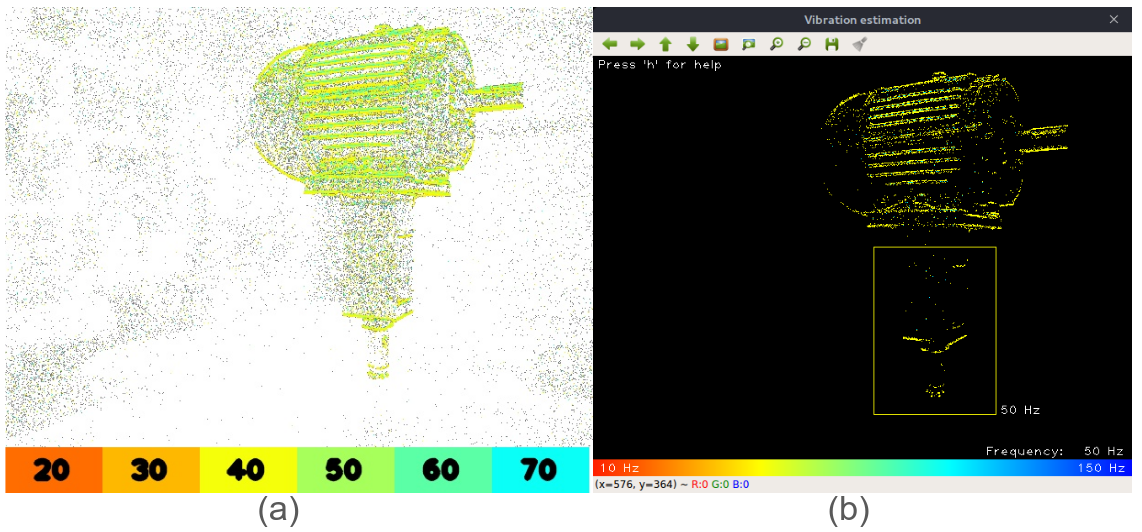}}
\caption{Our observation (a) gives similar result of 40-50 Hz frequency of vibrating motor as that of metavision vibration estimation (b). (b) is taken from \citep{propheseeVibrationEstimation}}
\label{fig:vibmotor}
\end{figure}

As in Fig:\ref{fig:vibmotor} Frequency map of a vibrating motor, with color range of frequencies (in Hz), captured by our algorithm gives frequency of vibrating motor to be in range of 40-50Hz which is compared to the closed source Prophesee’s metavision frequency analysis library \citep{propheseeVibrationEstimation}. Pixels depicted in grey represent areas where frequency detection was not estimated. 

\begin{figure}[h]
\centerline{\includegraphics[width=\linewidth]{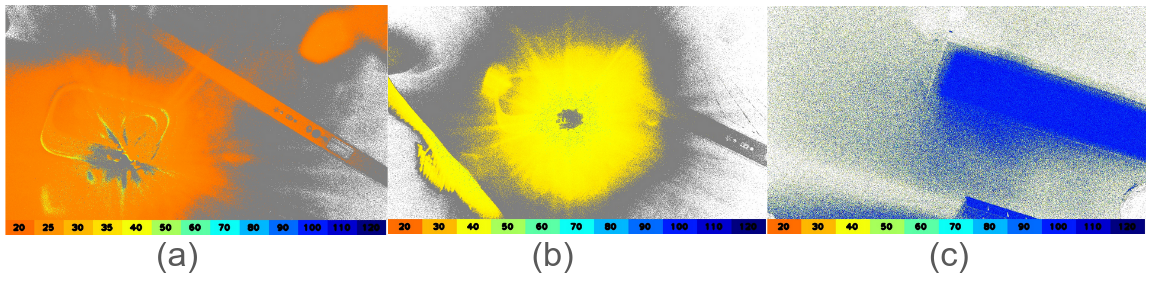}}
\caption{Capturing frequency map of light strobe from phone flasher (a and b) and ceiling led flicker (c) directly on event sensor at (a) 25 Hz rate (b) 40 Hz rate and (c) observed as 100 Hz and harmonic scatter of 50 Hz}
\label{fig:lightstrobe}
\end{figure}

During our experiments, we frequently observed LED flicker from ceiling lights in the event data, which needs to be filtered out. This can be achieved either by using the manufacturer's library to handle it at the hardware level during recording or by processing it at the software level afterward. To eliminate light flicker, it is important to identify which frequency band to filter, so visualizing the frequency map was helpful for estimating and removing flicker. We conducted a series of experiments with light strobes generated from a smartphone's LED flash in a dark room at specific frequencies—25, 40, 65, 85, 100, and 125 Hz, as illustrated in Fig. \ref{fig:lightstrobe} (a. and b.). Additionally, ceiling light flicker at 100 Hz is shown in Fig. \ref{fig:lightstrobe}. One application of detecting strobe light frequency, as discussed in \citep{6696456}, is to track and estimate pose based on specific strobe frequencies. Here to clearly capture flicker frequencies from structure we had to process event data in temporal batches of >45 ms. 

\begin{figure}[h]
\centerline{\includegraphics[width=\linewidth]{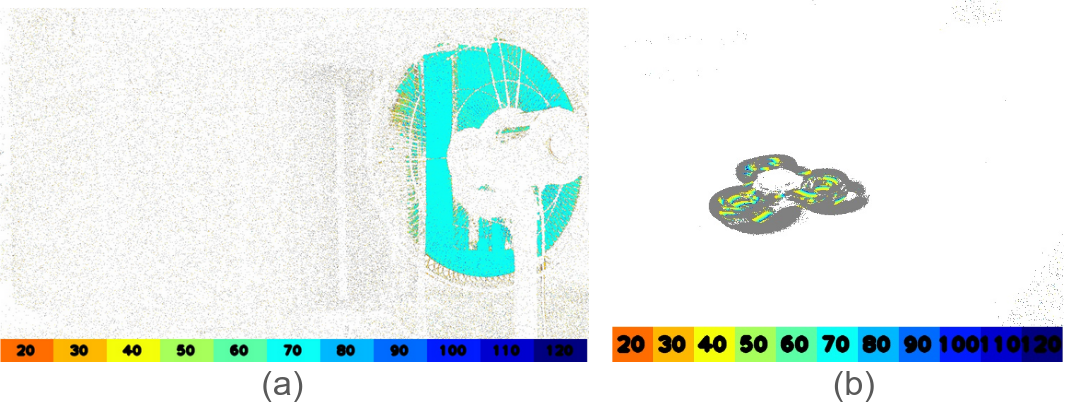}}
\caption{Shows frequency estimation of spinning objects like pedestal fan (a) and hand spinner (b).}
\label{fig:spinfreq}
\end{figure}

Frequency visualization of rotating objects offers critical insights into their dynamic behavior and operational performance. Techniques such as high-speed imaging and stroboscopic analysis are commonly employed to achieve this visualization. As illustrated in Fig:\ref{fig:spinfreq}, the use of frequency mapping enhances our ability to capture intricate details of motion dynamics of a pedestal fan and a hand spinner. This approach enables a more nuanced understanding of movement stability. Here to capture frequency of moving blades we had to process event data in temporal batches of 20 to 25 ms. 

\begin{figure}[h]
\centerline{\includegraphics[width=\linewidth]{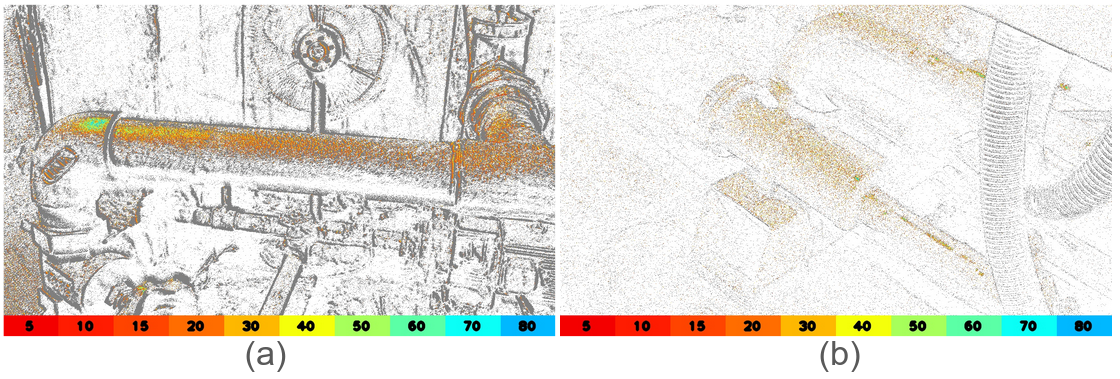}}
\caption{Shows frequency estimation of vibrating structures in water pumping station. (a.) shows low vibration (around 5-15 Hz) on below pipes and left most pipes and (b.) shows more vibration (around 20-30 Hz) as we localized more vibration noise from this section. Ignore 50hz ceiling led light flicker shinning in (a.)}
\label{fig:structfreq}
\end{figure}

Vibrating structures are a common occurrence in water pumping stations due to the rotating machinery, such as pumps, motors, and turbines, as well as from the flow of water through pipes and valves. While some level of vibration is normal and expected, excessive vibration can lead to various issues, including reduced efficiency, increased wear and tear, and potential structural damage.
\begin{itemize}
\item Early detection of problems: Regular vibration monitoring can help identify issues like misalignment, imbalance, or bearing wear before they escalate into major failures. This predictive maintenance approach can significantly reduce downtime and repair costs.
\item Safety improvements: Monitoring vibrations helps ensure that equipment is operating within safe parameters, reducing the risk of catastrophic failures that could pose safety hazards to personnel.
\item Performance optimization: Vibration analysis can provide insights into system performance, allowing operators to fine-tune operations for optimal efficiency and output.
\end{itemize}
To effectively measure vibrations, water pumping stations typically employ various sensors and monitoring systems, such as accelerometers, velocity sensors, and displacement probes \citep{6749391}. Such setups can have complex build and analysis complexity. These are helpful but are not sufficient to localize the issue precisely, with frequency image monitoring (see Fig:\ref{fig:structfreq}) we can observe on pixel by pixel level where exactly how much vibration is generated and what could be the source and further visualize that region for more understanding. Here to capture vibrations from structure we had to process event data in temporal batches of >15 ms. 

\subsection{Qualitative Evaluation of Motion Magnification}
In our exploration of this technique, we have focused on a specific approach that utilizes steerable pyramids \citep{537667}. To apply motion magnification to event data, its required to reconstruct intensity image from event data (see Section~\ref{sec:ReconstructBasics}) before applying steerable pyramid technique (see Section~\ref{sec:steerablepyr}).

\begin{figure}[h]
\centerline{\includegraphics[width=\linewidth]{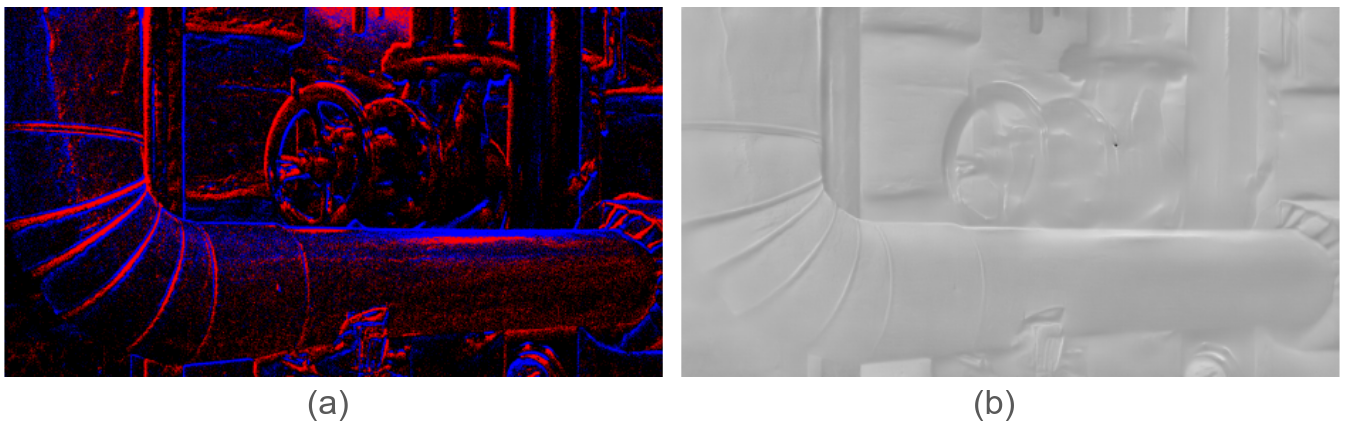}}
\caption{Shows event data in blue/red polarity (a) and its intensity reconstructed image (b).}
\label{fig:reconstructEvent}
\end{figure}

\noindent E2VID method's \citep{rebecq2019highspeedhighdynamic} data-driven approach represents a significant shift from conventional methods, potentially offering more robust and adaptable reconstruction capabilities across a wide range of visual scenarios. As you can see in Fig:\ref{fig:reconstructEvent}, (a) shows event data with both polarity and its intensity reconstructed image in (b) with the use of E2VID network.

\noindent Our observations with Intensity Reconstruction:
\begin{itemize}
\item Over Smoothening Issue in reconstruction: Existing models trained on synthetic data tend to produce undesired outputs when applied to scenarios where only a part of the scene moves.
\item Model Generalization in reconstruction: Current models fail to generalize well to scenarios with stationary cameras and moving scene components.
\end{itemize}

\noindent Further possible improvements for reconstruction:
\begin{itemize}
\item Custom Dataset Collection: Plan to capture real event data, focusing on scenarios with stationary cameras and moving scene elements.
\item Model Retraining: Utilize the dataset to retrain the intensity reconstruction model for better performance in scenarios with isolated motion.
\end{itemize}

\begin{figure}[h]
\centerline{\includegraphics[width=\linewidth]{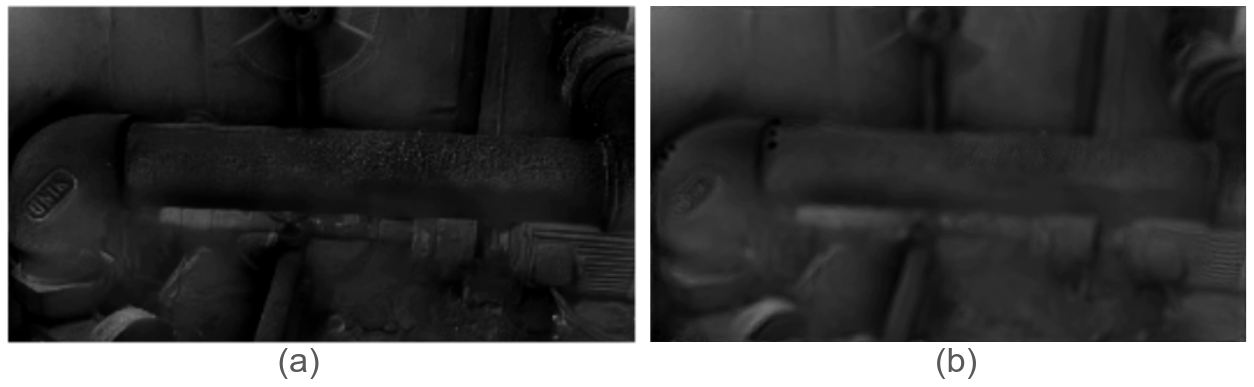}}
\caption{Shows intensity reconstructed image (a) and respective motion magnification frame (b).}
\label{fig:magEvent}
\end{figure}

\noindent Since, there is no current method for directly magnifying event data, we took the hybrid approach. We first reconstructed the intensity images from event data (E2VID)\citep{rebecq2019highspeedhighdynamic}, then using the Phase based approach magnified the video (see Fig:\ref{fig:magEvent}). For magnification, we required the frequency range of motion, which was easily known from the frequency map estimation mentioned earlier. 
\newline However, this had a major conflict.
\begin{itemize}
\item Camera Stability vs. Scene Motion: Motion magnification typically excels when the camera remains stable while specific parts of the scene exhibit motion.
\item Inherent Conflict: Intensity image reconstruction methods for event data operate optimally when the camera moves, while motion magnification demands stability for accurate interpretation of motion.
\end{itemize}

\noindent Further possible improvements for motion magnification:
\begin{itemize}
\item Instead of intensity reconstruction better way would be to directly use the per pixel frequency map, estimate phase based on intensity reconstruction and applying optical flow to enhance subtle motions.
\item To visualize the scene there is no other way than reconstruction but one can simply apply motion magnification on last reconstructed frame and show motion magnification for complete video stable on a scene.
\end{itemize}

\section{Conclusion}
\label{sec:conclusion}
This report has explored cutting-edge applications of event-based vision sensors, focusing on vibrational analysis and its extension of motion magnification. Our investigation revealed both the potential and challenges of these technologies in real-world scenarios. In frequency measurement and vibrational analysis, event-based sensors demonstrated promising capabilities for real-time structural assessment. It concludes over ON-to-OFF event change approach for frequency estimation method for jitter free stable per-pixel frequency estimation. However, our work with intensity reconstruction for the purpose of motion magnification highlighted significant challenges, including over-smoothing issues and limited model generalization, particularly in scenarios with stationary cameras and isolated motion. To address these limitations, we propose custom dataset collection and model retraining strategies. Our exploration of motion magnification techniques uncovered an inherent conflict between the optimal conditions for intensity image reconstruction and those for accurate motion magnification. This finding underscores the complexity of applying event-based vision to subtle motion analysis.

\noindent Looking forward, we suggest innovative approaches to overcome these challenges. These include utilizing per-pixel frequency maps and phase estimation based on intensity reconstruction, combined with optical flow techniques for enhancing subtle motions. While event-based vision sensors show great promise for advanced applications like vibrational analysis and motion magnification, significant work remains to fully realize their potential. Future research should focus on bridging the gap between theoretical capabilities and practical implementation, paving the way for more robust and versatile event-based vision systems in structural analysis and beyond.

\section{Acknowledgements}
The authors would like to acknowledge the support from CYRAN AI Solutions.

\bibliographystyle{unsrtnat}
\bibliography{references}  






\end{document}